\documentclass[10pt,twocolumn,letterpaper]{article}

\usepackage[pagenumbers]{cvpr} 

\usepackage{float}

\definecolor{cvprblue}{rgb}{0.21,0.49,0.74}
\usepackage[pagebackref,breaklinks,colorlinks,allcolors=cvprblue]{hyperref}

\newcommand{\name}{ModSkill\xspace}

\title{\name: Physical Character Skill Modularization}

\author{Yiming Huang$^{1}$, Zhiyang Dou$^{1,2}$, Lingjie Liu$^{1}$\\
University of Pennsylvania$^{1}$, The University of Hong Kong$^{2}$\\
{\tt\small \{ymhuang9, zydou, lingjie.liu\}@seas.upenn.edu}
}

\begin{document}
 \twocolumn[{
 \renewcommand\twocolumn[1][]{#1}
\maketitle
\begin{center}
    \vspace{-0.5cm}
    \captionsetup{type=figure}
     \includegraphics[width=\linewidth]{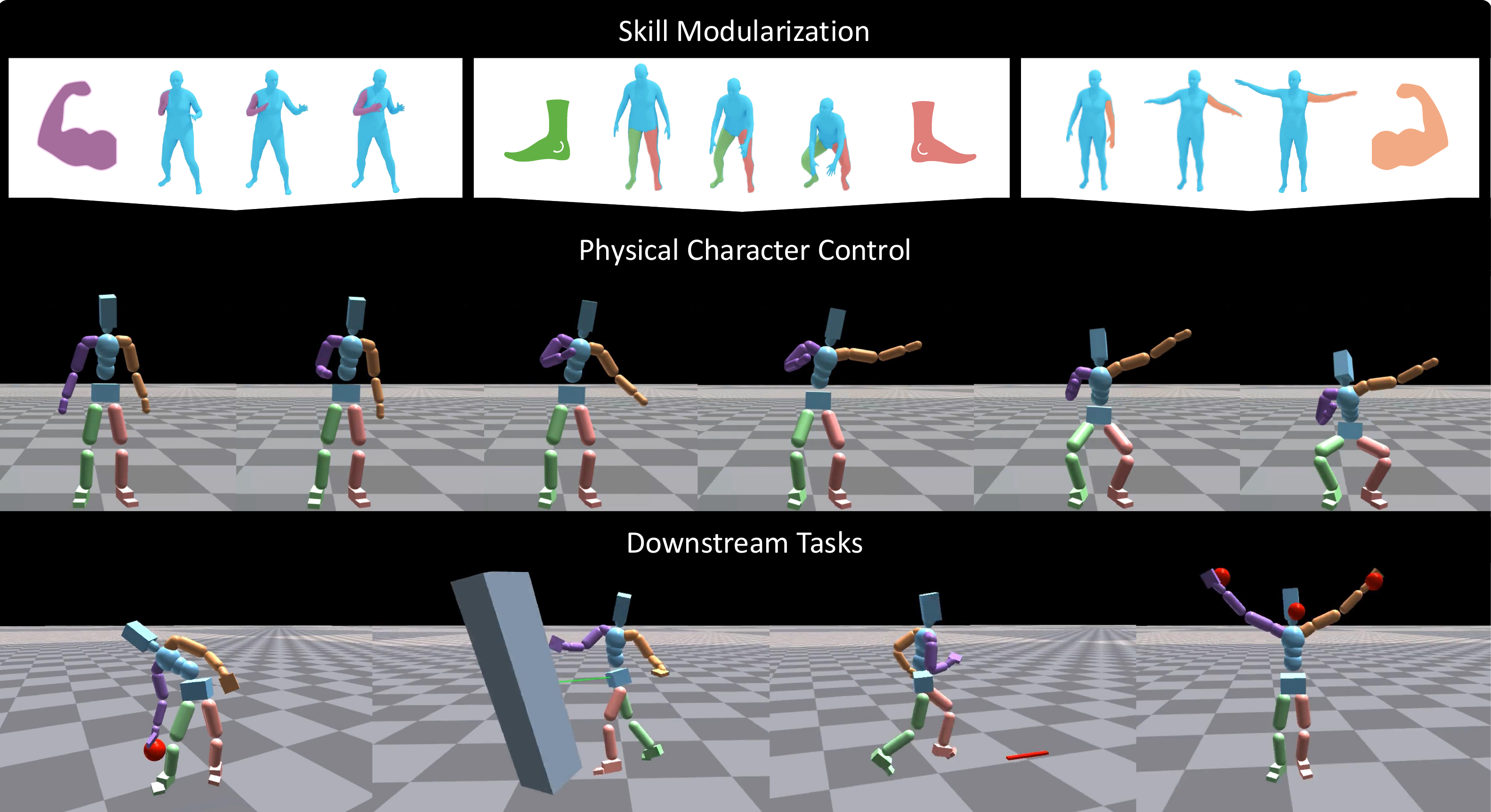}
    \captionof{figure}{We propose a modularized skill learning framework, \name, that decouples full-body motion into skill embeddings for controlling individual body parts. Learned from large-scale motion datasets, these modular skills can be combined to control a simulated character to perform diverse motions, such as the Usain Bolt pose, and seamlessly reused for various downstream tasks.}
    \label{fig:teaser}
\end{center}}
]
\begin{abstract}
Human motion is highly diverse and dynamic, posing challenges for imitation learning algorithms that aim to generalize motor skills for controlling simulated characters. Previous methods typically rely on a universal full-body controller for tracking reference motion (tracking-based model) or a unified full-body skill embedding space (skill embedding). However, these approaches often struggle to generalize and scale to larger motion datasets. In this work, we introduce a novel skill learning framework, \name, that decouples complex full-body skills into compositional, modular skills for independent body parts. Our framework features a skill modularization attention layer that processes policy observations into modular skill embeddings that guide low-level controllers for each body part. We also propose an Active Skill Learning approach with Generative Adaptive Sampling, using large motion generation models to adaptively enhance policy learning in challenging tracking scenarios. Our results show that this modularized skill learning framework, enhanced by generative sampling, outperforms existing methods in precise full-body motion tracking and enables reusable skill embeddings for diverse goal-driven tasks. Project Page: \url{https://yh2371.github.io/modskill/}.
\end{abstract}    
\section{Introduction}
\label{sec:intro}
Physically simulated characters are widely used in animation \cite{:10.2312/SCA/SCA12/221-230,Wang2024PacerPlus}, VR/AR \cite{Luo_2024_CVPR, 10.1145/3550469.3555411}, and robotic tasks \cite{he2024omnih2o,he2024learning}. When combined with large-scale human motion capture data \cite{AMASS:ICCV:2019}, imitation learning policies can enable such characters to imitate a variety of motion skills \cite{Luo2023PerpetualHC, luo2024universal, juravsky2024superpadl, tessler2024masked}. However, the inherent diversity of human motion presents significant generalization challenges for imitation learning, which can be prone to overfitting.

Previous approaches to physical character skill learning can be typically classified into two categories. \textit{Tracking-based methods} train controllers to imitate reference motions by tracking target pose sequences from motion clips. Recently, progressive learning techniques have been utilized to gradually extract more complex body-level motion skills from diverse data into a set of expert controllers \cite{Luo2023PerpetualHC, juravsky2024superpadl}. The motor skills learned from this mixture of experts can be distilled into a compact universal motion representation that offers broader coverage of human motion \cite{luo2024universal}. However, such methods still fall short in addressing scalability challenges with larger datasets, requiring more experts and increasing manual effort for skill extraction. On the other hand, \textit{Skill embedding methods} employ hierarchical frameworks that pre-train compact skill embedding spaces, which are then repurposed for high-level tasks guided by carefully designed, task-specific rewards \cite{dou2022case,2022-TOG-ASE,juravsky2022padl,tessler2023calm,zhang2023vid2player3d}. However, the expressivity limitations of these latent skill spaces hinder their ability to capture the diverse range of body-level human motion skills present in larger motion datasets.

In this work, we argue that human motion is inherently modular, as evidenced by neuroscience and evolutionary developmental biology~\cite{sylos2022complexity, kitano2002computational,hintze2008evolution,callebaut2005modularity}. Motivated by this modularity, we aim to achieve \textit{Skill Modularization} by decoupling full-body motion and focusing on the independent control of individual body parts. 
Compared to body-level skills, modular skills for independent body parts are not only more compact but also exhibit a compositional quality that can generate highly diverse full-body motion \cite{Jang_2022, 10.1145/3550454.3555489}. By emphasizing these compact part-level skill spaces, we can simplify skill learning and enhance policy performance. Building on this intuition,  we introduce \name, a novel modularized skill learning framework that utilizes a motion imitation objective to effectively decouple body-level motor skills in large-scale motion datasets \cite{AMASS:ICCV:2019} into reusable, part-specific skills that each guide an independent low-level controller. Specifically, our approach incorporates a skill modularization attention layer that analyzes the relationships between part-specific observations and generates spherical modular skill embeddings for each controller, directing the corresponding body parts of the simulated agent. 

Additionally, we propose an \textit{Active Skill Learning} scheme by introducing a Generative Adaptive Sampling strategy that leverages pre-trained large motion generation models \cite{tevet2023human} to produce new samples for challenging motion sequences with respect to each body part. In stark contrast to previous efforts~\cite{dou2022case, Luo2023PerpetualHC}, where resampled motion clips consistently come from a fixed motion dataset, our method applies a powerful generative model to provide prior-level resampling capabilities, thereby enhancing the skill learning process with more diverse motion samples. 

\name, with its modularity and compositionality, effectively learns diverse motor skills, enabling it to be reused by high-level policies for various downstream tasks. We conduct extensive experiments to demonstrate that our method achieves state-of-the-art performance both on full-body tracking-based tasks and across a wide range of generative, goal-driven benchmarks, including steering, reaching, striking, and VR tracking. In summary, our contributions are three-fold:

\begin{itemize} 
\item We propose a modularized skill learning framework that integrates a skill modularization attention layer for extracting part-specific skill embeddings that guide independent low-level controllers for each body part.
\item We introduce an Active Skill Learning scheme with a Generative Adaptive Sampling strategy that enhances motion imitation performance using large motion generation models. 
\item Our modular controllers achieve superior performance in precise motion tracking, and the learned part-wise skills are effectively reusable for generative downstream tasks. \end{itemize}

\begin{figure*}[t]
    \centering
    \includegraphics[width=\linewidth]{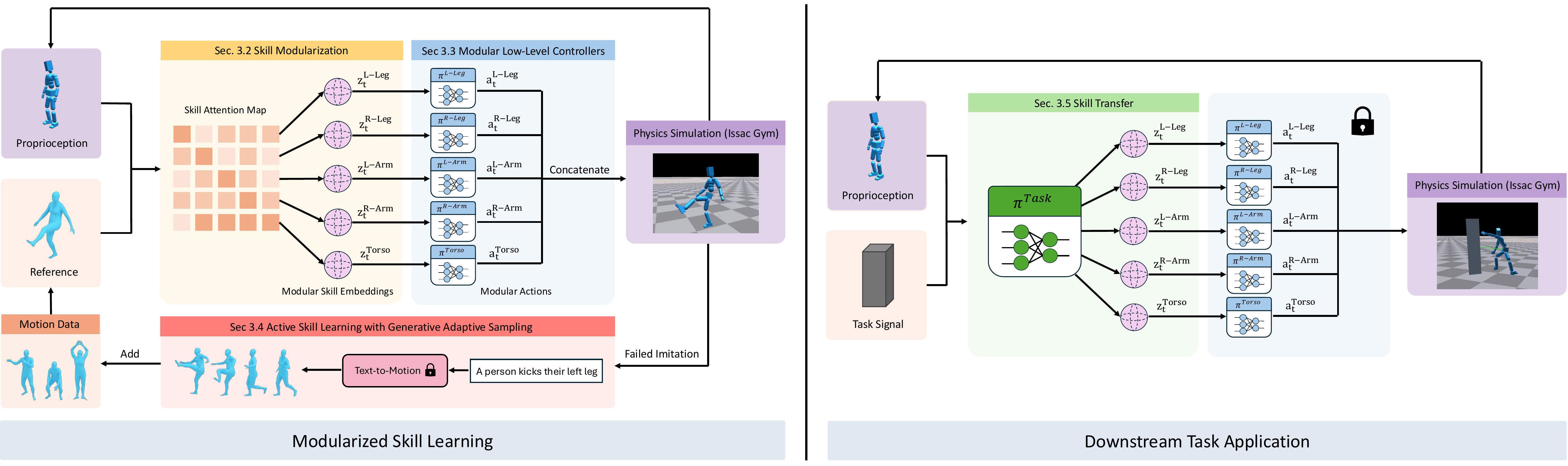}
    \caption{\textbf{Left}: We extract modular skills from a large-scale motion dataset using a motion imitation objective, enabling low-level controllers to control various body parts of a physically simulated character. Active skill learning, through adaptive sampling from an off-the-shelf motion generation model, further enhances policy performance. \textbf{Right}: The learned modular skills can be transferred to downstream tasks by freezing the low-level controllers and training a high-level policy with task-specific rewards.}
    \label{fig:Pipeline}
\end{figure*}

\section{Related Work}
\label{sec:related}

\textbf{Physics-based Motion Imitation.} Reproducing diverse and realistic human motions using physics-based characters has a long history and has gained significant attention recently~\cite{faloutsos2001composable, liu2010sampling, liu2005learning, liu2018learning, liu2017learning,
2018-TOG-deepMimic, 10.1145/3272127.3275014, 10.1145/3072959.3073602, zhang2023vid2player3d, 10.1145/3588432.3591525}. Due to the high diversity of human motion, many motion imitation approaches solely focus on task-specific scenarios that require only a subset of motor skills \cite{zhang2023vid2player3d, 10.1145/3588432.3591525, wang2024skillmimic, 10.1145/3450626.3459761, cui2024anyskill}. To generalize motion controllers across larger datasets, adversarial motion priors for motion tracking have been introduced \cite{Luo2023PerpetualHC, 10.1145/3450626.3459670}. Specifically, mixtures of experts have been widely employed, where each expert can focus on a specific atomic task, bridging the gap between task-specific motion imitation and general motion imitation \cite{10.5555/3454287.3454618, ScaDiver, Luo2023PerpetualHC, wagener2022mocapact, Luo2022FromUH}. However, the reliance on multiple experts introduces scalability challenges, as adding more experts may require increasing manual effort for more complex skill learning. Furthermore, previous methods predominantly focus on body-level skills, which may limit the expressiveness of the controller. In contrast, we address challenges in motion imitation by decoupling full-body motion into part-specific motor skills, leveraging modularization to simplify and improve skill learning.\\

\noindent \textbf{Physics-based Skill Embedding.} Adversarial learning and motion imitation provides a framework for developing reusable motion skills that can be applied across a variety of downstream tasks. In hierarchical learning settings, adversarial methods have been used to map structured latent embeddings to reusable low-level motor skills \cite{2022-TOG-ASE, tessler2023calm, dou2022case, juravsky2022padl}. While these skills transfer effectively to specialized high-level tasks trained on tailored motion data, they face challenges when scaling to more diverse motion. 
On the other hand, efforts have been made to model probabilistic latent spaces that can capture motor skills from larger datasets \cite{PhysicsVAE, merel2018neural, ling2020character, zhu2023NCP}. Moreover, prior work has shown that precise motion imitation policy networks can be distilled into universal motion representations, enabling coverage of large-scale motion datasets \cite{wagener2022mocapact, luo2024universal, juravsky2024superpadl}. In contrast to these approaches, our method focuses on decoupling motor skill learning during motion imitation by emphasizing the formulation of modular, part-level skills, allowing us to achieve effective and reusable skills without the need for additional distillation.\\

\noindent \textbf{Kinematics-based Motion Generation.} The rich latent space of  kinematics-based motion generation models  enables the generation of a wide variety of motion patterns from multi-modal condition signals \cite{jiang2024motiongpt, tevet2023human, chen2023executing, Guo_2022_CVPR, zhang2023generating}. Prior work has shown that synthetically generated data can provide valuable supervision for training and refining generative models \cite{gillman2024selfcorrecting, xie2023template_free, Black_CVPR_2023}. Furthermore, when integrated with a physics-based controller, motion generation models can effectively guide task planning and execution \cite{tevet2024closdclosingloopsimulation, yuan2023physdiff, ren2023insactor}. In this work, we propose to leverage the expressive power of large motion generation models to adaptively create synthetic examples of challenging motion imitation scenarios, enhancing policy learning and generalization.\\

\noindent \textbf{Part-level Motion Learning.} Previous work has demonstrated the advantages of body-part-level motion learning in both kinematics-based motion generation models \cite{huang2024como, wan2023tlcontrol, zhou2024emdm, zhang2023finemogen, Jang_2022} and physical controllers \cite{10.1145/3588432.3591487, https://doi.org/10.1111/cgf.15174, 10.1145/3550454.3555489}. Lee et al. \cite{10.1145/3550454.3555489} demonstrate the compositionality of body-part-level motion by introducing an assembler module that combines body-part motion signals from different sources before directing a single controller. PMP \cite{10.1145/3072959.3073602} adapts adversarial motion priors to individual body parts, enabling the extraction of specialized style rewards from diverse motion datasets. Similarly, Xu et al. \cite{composite} leverages multiple discriminators for training a control policy to imitate reference motions of various body parts from different sources. However, both methods still rely on a single controller for body-level control, limiting the decoupling of motor skills across body parts. PartwiseMPC \cite{https://doi.org/10.1111/cgf.15174} takes a step further by decoupling motion planning, allowing independent planning for body parts alongside whole-body planning, which improves generalization to unseen environments. In contrast, our approach introduces compact, modularized skill embeddings and low-level controllers for directing individual body parts, providing effective and reusable modular skills that can be applied to both precise motion imitation tasks and generative, goal-driven tasks.
\section{\name}
\label{sec:method}

In this paper, we present \name, a modularized framework for extracting body-part-level motor skills from large-scale motion datasets through imitation learning. As illustrated in Fig. \ref{fig:Pipeline}, our policy network consists of two key components for modularized skill learning: 1) a skill modularization attention layer (Sec. \ref{sec:attention}) that generates spherical embeddings to capture body-part-specific skills, and 2) a set of low-level skill-conditioned controllers (Sec. \ref{sec:controller}) that control the movement of individual body parts. To further improve policy performance, we introduce a generative adaptive sampling strategy that incorporates synthetic data from motion generation models into policy training (Sec. \ref{sec:generative}). Additionally, we show that the modular skills learned in our framework can be effectively transferred to downstream tasks via a high-level, task-specific policy (Sec. \ref{sec:transfer}).

\subsection{Preliminaries}
\label{sec:prelim}

Given a reference motion sequence of $ T $ frames, $ s^{r}_{1:T} $, our policy network, denoted as $\pi_{\text{\name}}$, is tasked to control a simulated humanoid agent to imitate the reference motion. 
We model the policy network as a Markov Decision Process (MDP), $ M = \langle S, A, T, R, \gamma \rangle $, where $ S, A, T, R, \gamma $ represent the state space, action space, transition dynamics, reward function, and discount factor, respectively.\\

\noindent \textbf{States and Actions.} In line with prior work \cite{Luo2023PerpetualHC}, we use a humanoid agent based on the SMPL kinematic model \cite{SMPL:2015}, which consists of 24 rigid bodies, 23 of which are actuated. The state $ s_{t} $ at time $ t $ is composed of two components: the proprioceptive state $ s^{p}_{t} $ and the reference state $ s^{r}_{t} $. The proprioceptive state $ s^{p}_{t} $ describes the current simulated configuration of the agent and is defined as:
\begin{equation}
    s^{p}_{t} := (\theta_{t}, p_{t}, v_{t}, \omega_{t})
\end{equation}
where $ \theta_{t} $, $ p_{t} $, $ v_{t} $, and $ \omega_{t} $ are the simulated joint rotations, positions, velocities, and angular velocities, respectively.

The reference state $ s^{r}_{t} $ encodes the target joint poses of the next time step, $t+1$, and the differences between the target joint poses and velocities of the next time step and the corresponding simulated values at the current time step, $t$. Specifically, it is defined as:
\begin{equation}
\small
    s^{r}_{t} := (\hat{\theta}_{t+1} \ominus \theta_t, \hat{p}_{t+1} - p_t, \hat{v}_{t+1} - v_t, \hat{\omega}_{t+1} - \omega_t, \hat{\theta}_{t+1}, \hat{p}_{t+1})
\end{equation}
where $ \ominus $ denotes the rotation difference. Here, $ \hat{\theta}_{t+1} $, $ \hat{p}_{t+1} $, $ \hat{v}_{t+1} $, and $ \hat{\omega}_{t+1} $ represent the reference joint rotations, positions, velocities, and angular velocities for the next time step, respectively. Both $ s^{r}_{t} $ and $ s^{p}_{t} $ are canonicalized with respect to the agent's local coordinate frame.

A PD (proportional-derivative) controller is applied at each joint, enabling the humanoid agent to follow the desired joint angles and velocities based on the reference motion sequence. The action space consists of PD control targets for each actuated joint, where the action directly specifies the PD targets without relying on residual forces \cite{yuan2020residual} or residual control \cite{Luo2022EmbodiedSH}.\\

\noindent \textbf{Reward.} We follow \cite{Luo2023PerpetualHC} to define the reward term as the sum of an imitation reward $r_{\text{imitation}}$, a discriminator reward $r_{\text{amp}}$ using the same setup as Adversarial Motion Prior \cite{10.1145/3450626.3459670} to encourage the combination of body-part-level motor skills to produce natural full-body motion, and an energy penalty reward  $r_{\text{energy}}$ to encourage smoother motion \cite{10.1145/3550469.3555411}:
\begin{equation}
    r := r_{\text{imitation}} + r_{\text{amp}} + r_{\text{energy}}
\end{equation}

\noindent Specifically, the imitation reward $r_{\text{imitation}}$ encourages the humanoid agent to imitate the reference motion by minimizing the difference between the translation $(p_{t}, \hat{p}_{t})$, rotation $(\theta_{t}, \hat{\theta}_{t})$, linear velocity $(v_{t}, \hat{v}_{t})$, and angular velocity $(\omega_{t}, \hat{\omega}_{t})$ of the simulated character and the target motion:
\begin{equation}
    \begin{aligned}
            r_{\text{imitation}} \ := w_{p}e^{-\lambda_{p}\|p_{t} - \hat{p}_{t}\|} + w_{\theta}e^{-\lambda_{\theta}\|\theta_{t} - \hat{\theta}_{t}\|} \\
    \ + w_{v} e^{-\lambda_{v}\|v_{t} - \hat{v}_{t}\|} + w_{\omega}e^{-\lambda_{\omega}\|\omega_{t} - \hat{\omega}_{t}\|}
    \end{aligned}
\end{equation}
where $ w_{\{ \cdot \}} $, $ \lambda_{\{ \cdot \}} $ denote the corresponding weights.

\begin{figure}[t]
    \centering
    \includegraphics[width=\linewidth]{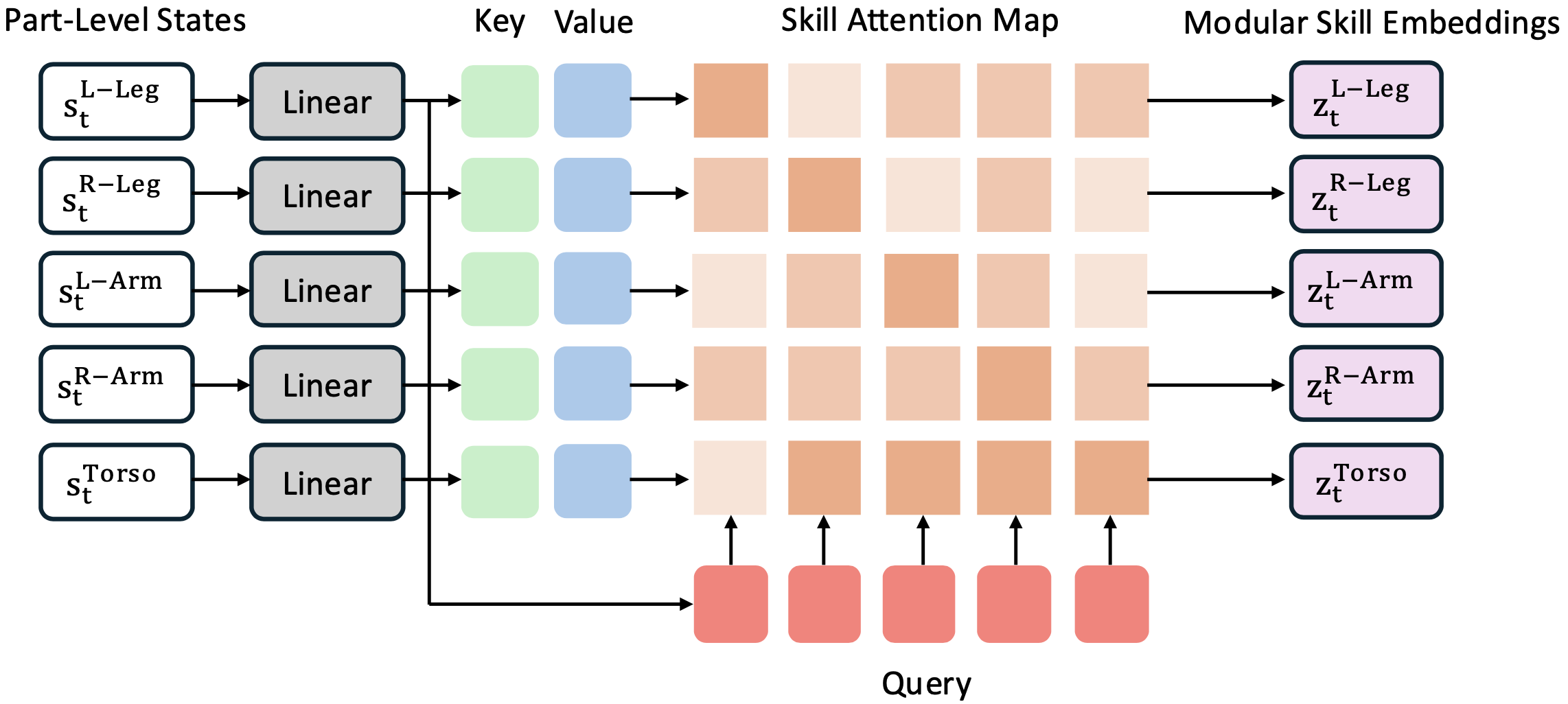}
    \vspace{-0.5cm}
    \caption{\textbf{Skill Modularization Attention Layer:} Given partial states for each body part, attention between body parts produces modular skill embeddings.}
    \label{fig:attention}
\end{figure}

\begin{figure*}[h]
    \centering
    \includegraphics[width=\linewidth]{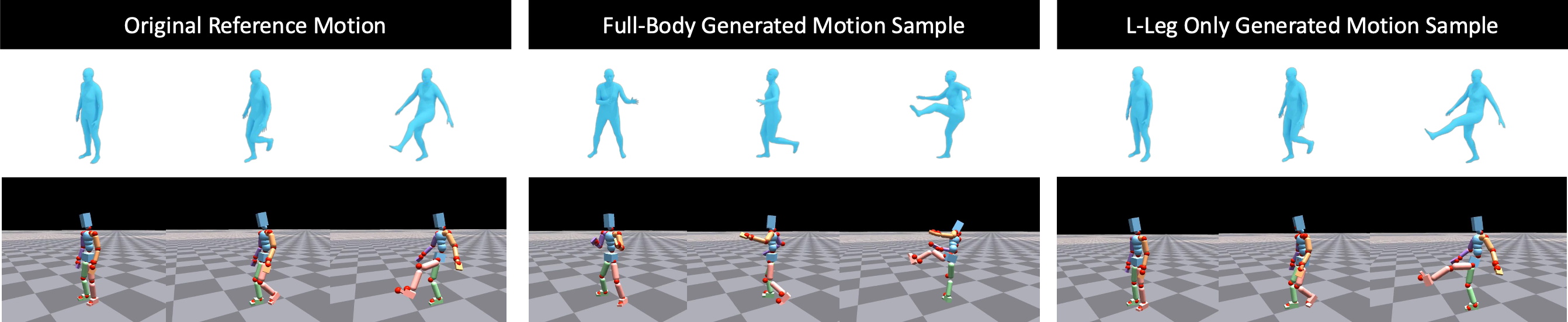}
    \caption{\textbf{Generative Adaptive Sampling}: When generating new samples for a reference motion (left), such as "a person kicks their left leg," the synthesized full-body motion (middle) introduces diverse variations from the original sequence. In contrast, synthesized motion for a specific body part (right) captures subtle differences, such as knee angles. The top row shows the target motion sequence, while the bottom row displays the imitated motion produced by our policy network. Red spheres indicate the corresponding target joint locations.}
    \label{fig:gensample}
\end{figure*}

\subsection{Skill Modularization}
\label{sec:attention}

To achieve modularized skill learning, we partition the rigid bodies of the simulated agent into $ K $ sets, each corresponding to a distinct body part. In this work, we set $ K = 5 $, corresponding to the set of body parts, $\mathcal{P} := $ \{Left Leg (L-Leg), Right Leg (R-Leg), Left Arm (L-Arm), Right Arm (R-Arm), Torso\}. Note that the specific partitioning is not restrictive, and can be adapted to suit different use cases or system configurations.

Let the state of all joints for body part $k \in \mathcal{P}$ at time $ t $ be denoted as $ s^{k}_{t} $. Our policy network incorporates an attention mechanism to enable high-level information sharing across body parts for effective skill modularization and encouraging whole-body consistency. As shown in Fig. \ref{fig:attention}, for each body part $ k \in \mathcal{P}$, the corresponding state $ s^{k}_{t} $ is projected into three vectors: key $ \mathbf{K}^{k}_{t} $, query $ \mathbf{Q}^{k}_{t} $, and value $ \mathbf{V}^{k}_{t} $. The attention scores between the query $ \mathbf{Q}^{k}_{t} $ of the current body part $k$ and the keys $ \mathbf{K}^{k'}_{t} $ from all body parts $ k' \in \mathcal{P}$ are computed by calculating the scaled dot-product between the query and each key,  and then passed through a softmax function to obtain the attention weights:
\begin{equation}
    \alpha_{t}^{k,k'} = \text{softmax}\left( \frac{\mathbf{Q}^{k}_{t} \cdot \mathbf{K}^{k'}_{t}}{\sqrt{d}} \right)
\end{equation}
where $ d $ is the dimension of the query and key vectors. The attention weights $ \alpha_{t}^{k,k'} $ indicate the relative importance of the states of different body parts when computing the skill embedding for body part $ k $.
The skill embedding $ z_{t}^{k} $ for each body part is then obtained by computing a weighted sum of the value vectors $ \mathbf{V}^{k'}_{t} $ from all body parts, with the attention weights serving as the coefficients:
\begin{equation}
    z_{t}^{k} = \sum_{k'} \alpha_{t}^{k,k'} \mathbf{V}^{k'}_{t}
\end{equation}
$ z_{t}^{k} $ is normalized with respect to $\|z_{t}^{k}\|$ to lie on the unit sphere. This normalization ensures that the skill embeddings are constrained within a consistent space, allowing for more stable learning \cite{2022-TOG-ASE}. By sharing information via this attention mechanism, each body part can focus on different aspects of the overall state for modularized skill learning.

\begin{table*}[h]
    \centering
    \caption{Full-body motion imitation results on AMASS train and test.}
    \begin{tabular}{@{}l|ccccc|ccccc@{}}
        \toprule
        & \multicolumn{5}{c|}{AMASS-Train} & \multicolumn{5}{c}{AMASS-Test} \\ 
        \midrule
        Method & Succ ↑ & $E_{\text{g-mpjpe}}$ ↓ & $E_{\text{mpjpe}}$ ↓ & $E_{\text{acc}}$ ↓ & $E_{\text{vel}}$ ↓ & Succ ↑ & $E_{\text{g-mpjpe}} $ ↓ & $E_{\text{mpjpe}}$↓ & $E_{\text{acc}}$ ↓ & $E_{\text{vel}}$ ↓ \\ 
        \midrule
        UHC & 97.0\% & 36.4 & 25.1 & 4.4 & 5.9 & 96.4\% & 50.0 & 31.2 & 9.7 & 12.1 \\ 
        PHC & 98.9\% & 37.5 & 26.9 & 3.3 & 4.9 & 97.1\% & 47.5 & 40.0 & 6.8 & 9.1 \\ 
        PULSE & 99.8\% & 39.2 & 35.0 & 3.1 & 5.2 & 97.1\% & 54.1 & 43.5 & 7.0 & 10.3 \\ 
        PHC+ & \textbf{100\%} & 26.1 & 21.1 & 2.6 & 3.9 & 99.2\% & 36.1 & 24.1 & 6.2 & 8.1 \\ 
        \name (Ours) & 99.6\% & \textbf{25.5} & \textbf{20.4} & \textbf{2.1} & \textbf{3.4} & \textbf{99.3\%} & \textbf{32.2} & \textbf{22.7} & \textbf{4.4} & \textbf{6.3} \\ 
        \bottomrule
    \end{tabular}
    \label{tab:track}
\end{table*}
\begin{table*}[h]
    \centering
    \caption{VR-tracking result on AMASS train and test.}
    \begin{tabular}{@{}l|ccccc|ccccc@{}}
        \toprule
        & \multicolumn{5}{c|}{AMASS-Train} & \multicolumn{5}{c}{AMASS-Test} \\ 
        \midrule
        Method & Succ ↑ & $E_{\text{g-mpjpe}}$ ↓ & $E_{\text{mpjpe}}$ ↓ & $E_{\text{acc}}$ ↓ & $E_{\text{vel}}$ ↓ & Succ ↑ & $E_{\text{g-mpjpe}}$ ↓ & $E_{\text{mpjpe}}$ ↓ & $E_{\text{acc}}$ ↓ & $E_{\text{vel}}$ ↓ \\ 
        \midrule
        ASE & 18.6\% & 128.7 & 87.9 & 40.9 & 33.3 & 8.0\% & 114.3 & 99.2 & 57.7 & 44.0 \\ 
        ASE-PMP & 7.2\% & 159.7 & 155.7 & 142.2 & 123.0 & 1.5\% & 161.7 & 126.1 & 151.1 & 96.4 \\ 
        PULSE & \textbf{99.5\%} & 57.8 & 51.0 & 3.9 & 7.1 & \textbf{93.4\%} & 88.6 & 67.1 & 9.1 & 14.9 \\ 
        \name (Ours) & 99.3\% & \textbf{52.9} & \textbf{47.9} &  \textbf{3.7} & \textbf{6.4} & \textbf{93.4\%} & \textbf{83.2} & \textbf{65.7} & \textbf{8.8} & \textbf{13.4} \\ 
        \bottomrule
    \end{tabular}
    \label{tab:vr}
\end{table*}
\subsection{Modular Low-Level Controllers}
\label{sec:controller}

To further enhance the flexibility and effectiveness of our policy network, we implement modular low-level controllers that operate alongside the skill modularization attention mechanism. For each body part $k$, we designate a low-level controller $\pi^{k} = \mathcal{N}(\mu(z_{t}^{k}), \sigma)$, which models a Gaussian distribution with fixed diagonal covariance. The skill embeddings, $z_t^{k}$, generated by the attention mechanism serve as the input for these controllers, which process the information to produce targeted actions $a_t^{k}$ for the actuated rigid bodies corresponding to body part $k$. The produced actions for each controller are concatenated to form the full-body PD target, denoted as $a_{t}$, for controlling the simulated agent. Using the same setup in \cite{10.1145/3450626.3459670}, we incorporate a body-level discriminator $D(s^{p}_{t-10:t})$ that computes a real/fake value based on the current body-level proprioception of the humanoid. This style signal encourages the formulation of natural body-level motion from modular part-level skills. By decoupling the action prediction into specialized modules, we enhance the flexibility of the network, enabling effective imitation for a wide range of motions.

\subsection{Active Skill Learning with Generative Adaptive Sampling}
\label{sec:generative}

Active skill learning enables efficient policy training by directing the learning process toward the most informative regions of the skill space. Previous methods \cite{Luo2023PerpetualHC, luo2024universal} often use adaptive sampling to prioritize motion sequences within the same fixed training set based on their failure probability for better motion imitation, which may lead to overfitting to challenging sequences. In contrast, we propose a generative adaptive sampling strategy that synthesizes $N$ motion sequences for each failed sample. These synthetic sequences are generated using an off-the-shelf text-to-motion diffusion model \cite{tevet2023human}, conditioned on the paired HumanML3D text descriptions \cite{Guo_2022_CVPR} of the failed motion. To better facilitate modularized skill learning, we not only generate new full-body samples but also leverage the controllability of the motion diffusion model to tailor new motion samples for each body part. In this process, we fix the reference motion for other body parts and allow the diffusion model to synthesize new motion samples for the target body part given the text condition (See Fig. \ref{fig:gensample}). Integrating these synthetic samples into the training process strengthens the policy network by providing a more balanced and comprehensive dataset. 

\subsection{Skill Transfer for Downstream Tasks}
\label{sec:transfer}

After $\pi_{\text{\name}}$ converges, a high-level policy $\pi_{\text{Task}}(z_{t}^{k_{1}}, ..., z_{t}^{k_{K}} | s_{t}^{p}, s_{t}^{g})$ can be trained to apply the learned modular skills to downstream tasks, where $s_{t}^{p}$ and $s_{t}^{g}$ represent the proprioceptive state and task-specific goal signal, respectively, and $z_{t}^{k_{1}}, ..., z_{t}^{k_{K}}$ indicate the corresponding spherical skill embeddings for each body part $k_{1}, ..., k_{K} \in \mathcal{P}$. Similar to the low-level controllers, each high-level task policy is modeled as a Gaussian distribution with a fixed diagonal covariance: $\mathcal{N}(\mu_{\text{Task}}(s_{t}^{p}, s_{t}^{g}), \sigma_{\text{Task}})$. When training the high-level policy, we freeze the low-level controllers to preserve the learned modular motor skills. In this work, we demonstrate the effectiveness and reusability of our modular part-level skills on a set of generative motion tasks. Details regarding the setup of each task are provided in the supplementary material.
\section{Evaluation}
\label{sec:eval}
\textbf{Experiment Settings.} \textit{Motion Tracking Task:} We evaluate \name's performance on the full-body motion tracking task, comparing it against state-of-the-art motion trackers UHC \cite{Luo2021DynamicsRegulatedKP}, PHC \cite{Luo2023PerpetualHC}, PHC+ and PULSE \cite{luo2024universal}. \\

\noindent \textit{Motion Skill Embedding Task:} We also highlight the reusability of our modular skill embeddings by applying \name to generative tasks such as reaching, steering, striking, and VR tracking, and comparing its performance with reusable skill representations: PULSE \cite{luo2024universal} and ASE \cite{2022-TOG-ASE}. Following \cite{luo2024universal}, we adapt ASE to produce per-frame latent skill embeddings for a fair comparison. Notably, while the original ASE uses a body-level adversarial motion prior, we will also compare the impact of partwise adversarial motion priors \cite{10.1145/3588432.3591487} on learning skill embeddings and high-level tasks (ASE-PMP). We adopt the same body part partition as our framework to formulate ASE-PMP.\\

\noindent \textbf{Datasets.} For training and testing the full-body and VR tracking policies, we utilize the cleaned AMASS training set and test set, respectively \cite{Luo2023PerpetualHC}. For the strike and reach tasks, we sample initial states from the AMASS training set. For speed tasks, we follow \cite{Wang2024PacerPlus} to use a subset of AMASS of only locomotion for initial state sampling.\\

\noindent \textbf{Metrics.} For motion imitation and VR controller tracking, we report the global end-effector mean per-joint position error ($E_{\text{g-mpjpe}}$) and root-relative end-effector mean per-joint position error ($E_{\text{mpjpe}}$) in millimeters. We also compare physics-based metrics, including acceleration error ($E_{\text{acc}}$) in mm/frame$^{2}$ and velocity error ($E_{\text{vel}}$) in mm/frame. Following prior work \cite{Luo2023PerpetualHC}, we define the success rate (Succ) as the percentage of time the average per-joint error remains within 0.5 meters of the reference motion. For VR tracking, the success rate is based on only three body points (Head, Left Hand, Right Hand). For generative tasks (reach, steer, strike), we compare the undiscounted return normalized by the maximum possible reward per episode. \\

\begin{figure*}[h]
    \centering
    \includegraphics[width=\linewidth]
    {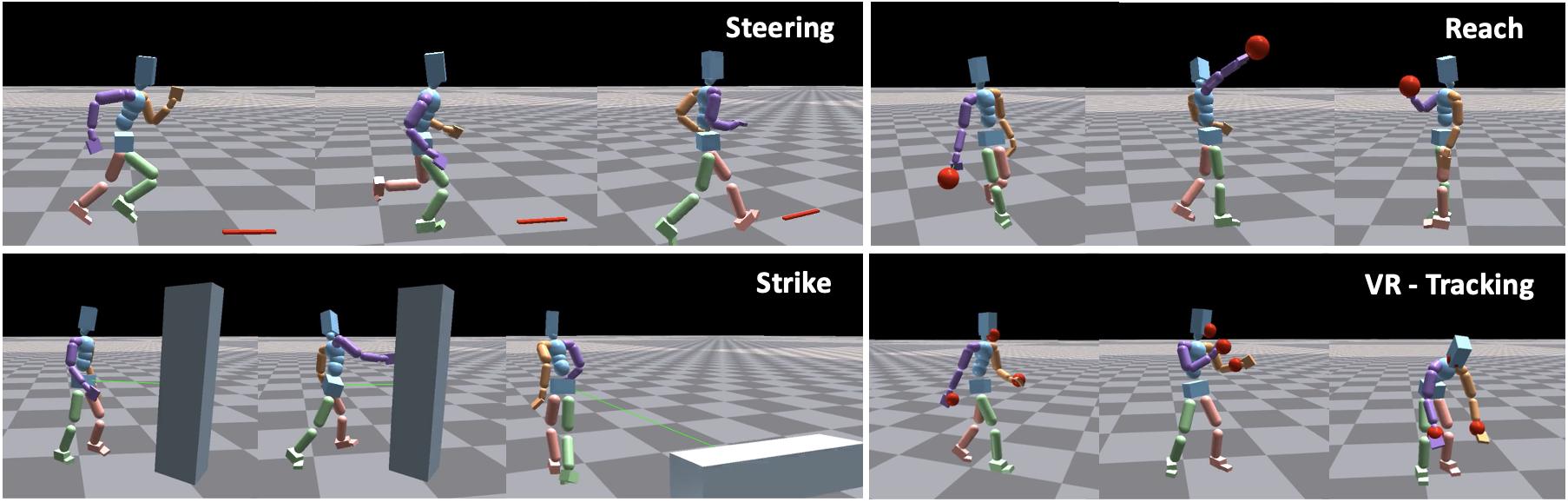}
    \caption{Our modular skill embeddings are flexible and informative, achieving natural human-like behavior in downstream tasks.}
    \label{fig:generative_qual}
\end{figure*}

\begin{table*}[h]
    \centering
    \caption{\textbf{Ablation on components of \name.} We evaluate on AMASS-Test for the full-body motion tracking task to demonstrate the effectiveness of each component for motor skill learning. Modularization: whether to use low-level controllers for each body part, Attention: whether to use a skill modularization attention layer. Generative: whether to use generative adaptive sampling. }
    \begin{tabular}{@{}cccc|cccccccc@{}}
        \toprule
        Index & Modularization & Attention & Generative &  Succ ↑ & \textbf{$E_{\text{g-mpjpe}}$ ↓} & \textbf{$E_{\text{mpjpe}}$ ↓} & \textbf{$E_{\text{acc}}$ ↓} & \textbf{$E_{\text{vel}}$ ↓} \\ 
        \midrule
        1 & $\times$ & $\times$ & $\times$  & 96.4\% & 41.1 & 28.4 & 5.4 & 7.2 \\ 
        2 & $\checkmark$ & $\times$ & $\times$  & 98.6\% & 35.9 & 23.8 & 4.4 & 6.5 \\ 
        3 & $\checkmark$ & $\checkmark$ &  $\times$ & 99.3\% & 32.4 & 23.2 & 4.5 & \textbf{6.3} \\ 
        \midrule
        4 & $\checkmark$ & $\checkmark$ & $\checkmark$ & \textbf{99.3\%} & \textbf{32.2} & \textbf{22.7} & \textbf{4.4} & \textbf{6.3} \\ 
        \bottomrule
    \end{tabular}
    \label{tab:ablation}
\end{table*}

\noindent \textbf{Implementation Details.} All physics simulations are conducted using Isaac Gym \cite{Makoviychuk2021IsaacGH}. Our policy network is trained with 3072 parallel environments on a single NVIDIA A6000 GPU for approximately one week, totaling around 2e9 steps. For PULSE, we use the original model settings. For \name and the ASE baselines, all low-level controller networks are four-layer perceptrons (MLPs) with dimensions [2048, 1536, 1024, 512]. Discriminators and encoders for adversarial skill learning are two-layer MLPs with dimensions [1024, 512]. Each high-level policy for downstream tasks is a three-layer MLP with dimensions [2048, 1024, 512]. The latent dimension of the skill embeddings is set to 64. Detailed hyperparameter settings are provided in the supplementary material. The controllers operate at 30 Hz, while the simulation runs at 60 Hz. 

\subsection{Motion Tracking}

Table \ref{tab:track} and Table \ref{tab:vr} show the performance of our method on the AMASS train and test sets for the full-body motion tracking task and VR tracking task, respectively. For both tracking tasks, our modular policy network outperforms baselines, reducing tracking errors on both training and test motion sequences. Our experiments show that by modularizing skills, a single network can achieve better accuracy and generalization capabilities. The results support the hypothesis that modularization of motor skills can effectively capture a wide range of human motion.

\begin{figure*}[h]
    \centering
    \includegraphics[width=\linewidth]{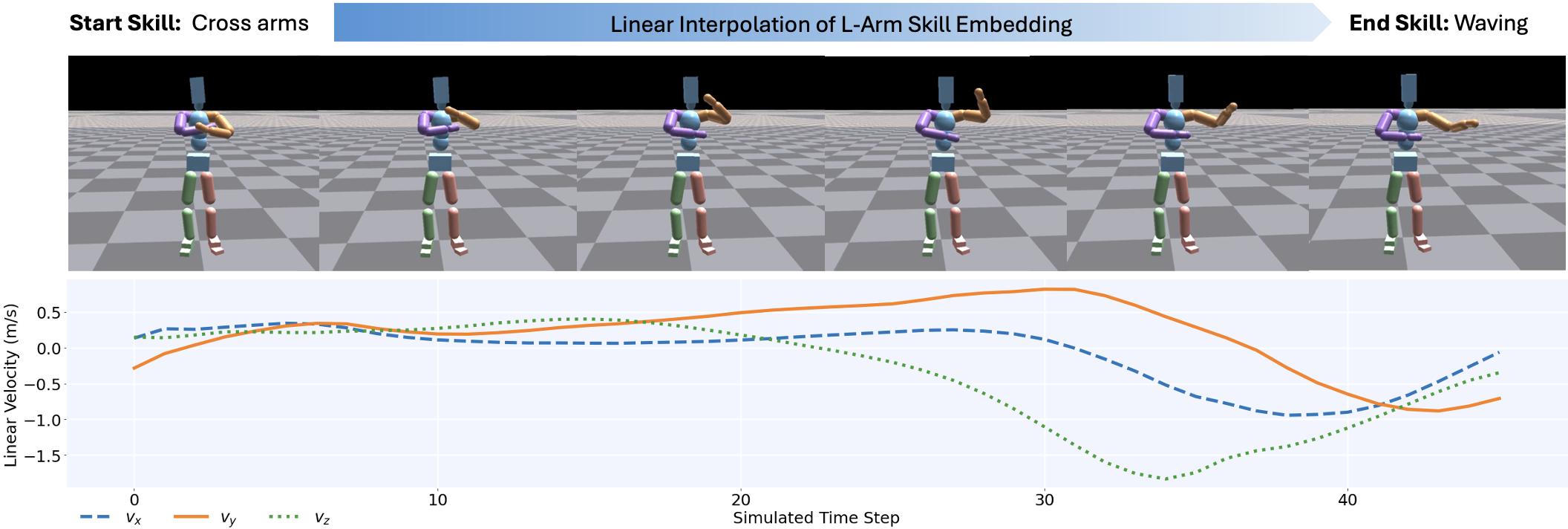}
    \caption{\textbf{Skill Interpolation}: Interpolating left-hand skill embeddings from "crossing arms" to "waving" shows a smooth transition in motion (top row) and linear velocity expressed in the local coordinate frame (bottom row) for the left-hand of the simulated character.}
    \label{fig:interpolate}
\end{figure*}

\subsection{Skill Embedding Downstream Tasks}
As shown in Fig. \ref{fig:generative_qual}, the modular skills learned by our framework can be effectively applied to downstream tasks. In Table \ref{tab:comparison}, we record the normalized return for the downstream tasks, steering, reach and strike, with 0 being the minimum possible return value, and 1 being the maximum. Compared to ASE baselines, our method achieves a more expressive skill space, resulting in superior normalized returns. When compared to the state-of-the-art model PULSE, our approach achieves comparable performance in terms of normalized return. Unlike PULSE, our policy network does not require additional distillation from a motion tracking policy to obtain effective skill embeddings. Instead, the learned modular skills can be directly applied to a variety of downstream tasks while preserving accurate motion imitation capabilities. In contrast, PULSE suffers from a significant decrease in motion tracking accuracy compared to the motion tracking policy, PHC+, used for distillation (see Tab.~\ref{tab:track}). For tasks that rely on full-body template movements, such as speed and strike, our method initially exhibits slower progress due to the more complex search space of modular skills, which requires not only skill search but also learning feasible compositions of modular skills. ASE-PMP shows similar trends, where part-level reward signals lead to a more complex skill learning process. Additionally, when PMP is applied to a single large-scale dataset rather than multiple small-scale specialized datasets of task-specific scenarios, this can lead to less effective skill learning and transfer. For more precise tasks, like reaching, which targets specific body parts, our method demonstrates faster learning, indicating better flexibility. We include training curves for downstream tasks in the supplementary material.

\begin{table}[h]
    \centering
    \caption{Normalized returns for downstream tasks: steering, reach, and strike. Values in parentheses indicate standard deviation.}
    \begin{tabular}{@{}l|ccc@{}}
    \toprule
    & Steering & Reach & Strike \\ \midrule
    ASE & 0.60 (0.001) & 0.10 (0.003) & 0.12 (0.006) \\
    ASE-PMP & 0.31 (0.002) & 0.06 (0.002) & 0.13 (0.004) \\
    PULSE & 0.92 (0.002) & 0.77 (0.002) & \textbf{0.88 (0.002)} \\
    \name & \textbf{0.93 (0.001)} & \textbf{0.79 (0.002)} & \textbf{0.88 (0.003)} \\ 
    \bottomrule
    \end{tabular}
    \label{tab:comparison}
\end{table}

\subsection{Skill Interpolation and Composition}

\begin{figure}[htbp]
    \centering
    \includegraphics[width=\linewidth]{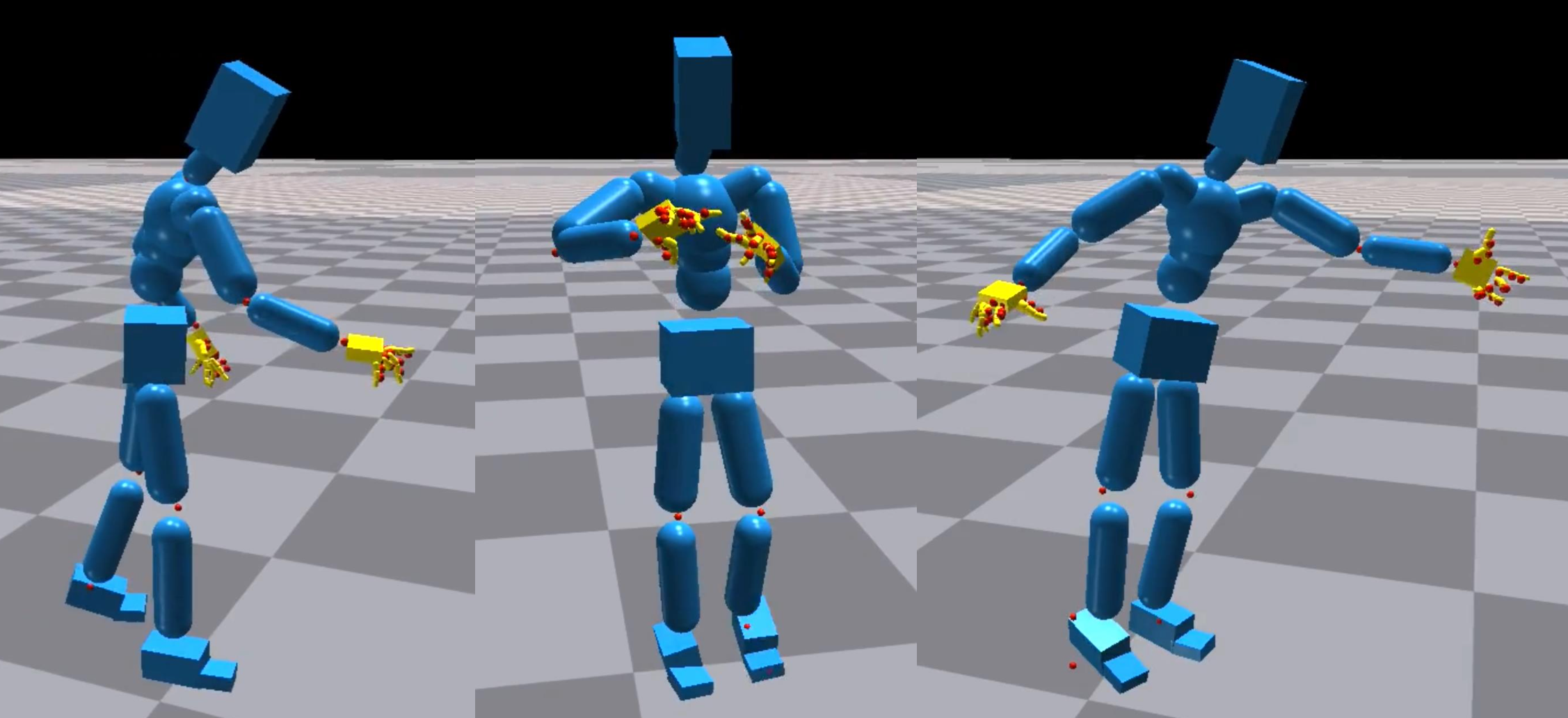}
    \caption{\name enables flexible integration of additional modular controllers without needing to retrain from scratch. By combining articulated hand controllers (highlighted in yellow) with our pre-trained body part controllers, we can control a humanoid with dexterous hands. Red spheres represent target joint locations.}
    \label{fig:hand}
\end{figure}

We demonstrate the structure of modular skill embeddings through interpolation. As shown in Fig. \ref{fig:interpolate}, we begin with the sequence of skill embeddings that controls the simulated character to cross both arms in front of its chest, and then gradually interpolate the left-hand skill embeddings towards those corresponding to a left-hand waving motion, while keeping the sequence of skill embeddings for all other body parts unchanged. The resulting motion exhibits a smooth transition, and further analysis of the left hand's linear velocity in the local coordinate frame over simulated time steps confirms this smoothness. As shown in Fig \ref{fig:teaser}, modular skill embeddings enable us to compose skills for different body parts, generating coordinated full-body motion, such as the Usain Bolt trademark pose. Due to this modularity, our framework also allows for flexible integration of new modular components. For example, as shown in  Fig. \ref{fig:hand}, we can combine articulated hand controllers with our pre-trained body part controllers to control a humanoid with dexterous hands to imitate a motion sequence without retraining the entire framework from scratch.

\subsection{Ablations}

We ablate the effectiveness of each component within our framework with respect to the full-body motion tracking task. In Table \ref{tab:ablation}, we present motion tracking results on the AMASS test set with respect to the following configurations: with/without modular controllers for individual body parts, with/without a skill modularization attention layer to extract modular skill embeddings, and with/without generative adaptive sampling. We observe that incorporating all components yields the best performance. Notably, even without the attention layer, our model outperforms PHC+ in motion tracking on unseen motion sequences, highlighting the effectiveness of modular motor skills for individual body parts. The addition of the attention layer not only improves performance but also enables the formulation of reusable modular motor skills for applications to downstream tasks. Furthermore, generative adaptive sampling enhances policy performance on unseen motions.
\section{Conclusion}
\label{sec:conclusion}

In this paper, we introduce a novel modularized skill learning framework, \name, which decouples complex full-body skills into compositional, modular skills for independent body parts. The framework incorporates a Skill Modularization Attention layer that transforms policy observations into modular skill embeddings, guiding independent low-level controllers for each body part. Additionally, our Active Skill Learning approach with Generative Adaptive Sampling utilizes large motion generation models to adaptively enhance policy learning in challenging tracking scenarios. Our results demonstrate that this modularized framework, enhanced by generative sampling, outperforms existing methods in achieving precise full-body motion tracking and enables reusable skill embeddings for diverse, goal-driven tasks.\\

\noindent \textbf{Limitations.} While modularized skill learning offers advantages in creating compact, compositional skill spaces, it introduces an inherent tradeoff in search complexity due to the formulation of multiple modular components. Future work could explore methods for improving the efficiency of modular skill composition. Additionally, incorporating environment-aware interaction priors would be a promising direction for extending the framework to interaction tasks.

{
    \small
    \bibliographystyle{ieeenat_fullname}
    \bibliography{main}
}

\clearpage
\setcounter{page}{1}
\setcounter{section}{0}
\maketitlesupplementary

\noindent This supplementary document provides additional implementation details (Sec.~\ref{sec:details}), an analysis of joint-level tracking errors comparing our method with PHC+ \cite{luo2024universal} (Sec.~\ref{sec:error}), task specifications along with training curves for downstream tasks (Sec.~\ref{sec:gen}), and visualizations of the modular skill embedding space (Sec.~\ref{sec:viz}). Extensive qualitative results are available in the supplementary video, where we demonstrate our method's capability to imitate various reference motion data and perform different downstream tasks.

\section{Implementation Details}
\label{sec:details}

In \name, we divide the 24 rigid bodies of the simulated character into five body parts, $\mathcal{P} := $ \{Left Leg (L-Leg), Right Leg (R-Leg), Left Arm (L-Arm), Right Arm (R-Arm), Torso\}. The corresponding grouping of rigid bodies is detailed in Tab. \ref{tab:group} 

\begin{table}[h]
    \centering
    \small
    \caption{Body part grouping for skill modularization.}
    \begin{tabular}{@{}l|l@{}}
        \toprule
        Body Part & Rigid Bodies \\ 
        \midrule
        L-Leg & L-Hip, L-Knee, L-Ankle, L-Toe\\
        R-Leg & R-Hip, R-Knee, R-Ankle, R-Toe\\
        Torso & Pelvis, Torso, Spine, Chest, Neck, Head\\
        L-Arm & L-Thorax, L-Shoulder, L-Elbow, L-Wrist, L-Hand\\
        R-Arm & R-Thorax, R-Shoulder, R-Elbow, R-Wrist, R-Hand\\
        \bottomrule
    \end{tabular}
    \label{tab:group}
\end{table}

\noindent Based on the body part grouping, we partition the input state accordingly. The Skill Modularization Attention Layer uses two-layer MLPs with dimensions [256, 64] to project the partitioned input states for each body part into keys, queries, and values, enabling attention across body parts. The resulting skill embeddings, with a dimensionality of 64, are then normalized to lie on the unit sphere. For each body part, we assign a low-level controller, implemented as a four-layer MLP with dimensions [2048, 1536, 1024, 512], which takes the corresponding skill embedding as input and produces PD targets for the actuated rigid bodies within that body part grouping. The policy network is trained for approximately $2 \times 10^9$ steps, with a learning rate of $2 \times 10^{-5}$. Generative adaptive sampling is applied every $2 \times 10^8$ steps. For each failed sequence, we utilize the corresponding text-label from HumanML3D \cite{Guo_2022_CVPR} to generate $N=3$ synthetic sequences using an off-the-shelf text-to-motion model \cite{tevet2023human}.\\

\noindent Note that \name is a skill-learning framework designed to be flexible with arbitrary body-part configurations. We adopted a 5-part setting that allows for clear separation between limbs and is consistent with prior work on part-based modeling \cite{Jang_2022}. In Tab. \ref{tab:bodypart}, we show additional motion tracking results on AMASS-Test for two-part (Part 1: L-Arm, R-Arm, Torso; Part 2: L-Leg, R-Leg) and three-part (Part 1: L-Arm, R-Arm; Part 2: Torso; Part 3: L-Leg, R-Leg) configurations. We also detail a comparison of the corresponding model sizes for different body part configurations using \name against
PHC+ \cite{luo2024universal}. Despite the smaller model size compared to PHC+, our five-part modularization approach leads to more precise motion tracking and enables the formulation of reusable, modular skills without the need for progressive mining. Additionally, for the two-part and three-part configurations, our framework demonstrates smaller model sizes than the full-body baseline while achieving competitive performance, highlighting the advantages of skill modularization.
\begin{table*}[h]
    \centering  
    \caption{Motion Tracking Evaluation and Policy Network Model Size for Different Body-Part Configurations.}
    \begin{tabular}{@{}l|ccccc|c@{}}
        \toprule
        Method & Succ ↑ & $E_{\text{g-mpjpe}}$ ↓ & $E_{\text{mpjpe}}$ ↓ & $E_{\text{acc}}$ ↓ & $E_{\text{vel}}$ ↓ & Model Size (MB)\\
        \midrule
        PHC+ & 99.2\%& 36.1 & 24.1 & 6.2 & 8.1 & $\sim$609\\
        \midrule
        2-Part & 97.1\% & 36.4 & 25.9 & 5.0 & 7.2 & $\sim$243\\
        3-Part & 98.6\% & 36.1 & 25.3 & 4.8 & 6.9 & $\sim$306 \\ 
        \name (Ours) & \textbf{99.3\%} & \textbf{32.2} & \textbf{22.7} & \textbf{4.4} & \textbf{6.3} & $\sim$429\\
        \bottomrule
    \end{tabular}
    \label{tab:bodypart}
\end{table*}



\begin{table}[h]
    \centering
    \footnotesize
    \caption{\textbf{R-Arm:} Joint-wise Mean Position Errors (mm) on AMASS-Test.}
    \begin{tabular}{@{}l|ccccc@{}}
        \toprule
        Method & R-Thorax & R-Shoulder & R-Elbow & R-Wrist & R-Hand \\ 
        \midrule
        PHC+ & 42.8 & 41.4 & 37.0 & 40.1 & 39.0\\
        \name & \textbf{20.6} & \textbf{21.2} & \textbf{21.8} & \textbf{29.8} & \textbf{30.5}\\
        \bottomrule
    \end{tabular}
    \label{tab:right_upper_limb}
\end{table}

\begin{table}[h]
    \centering
    \footnotesize
    \caption{\textbf{L-Arm:} Joint-wise Mean Position Errors (mm) on AMASS-Test.}
    \begin{tabular}{@{}l|ccccc@{}}
        \toprule
        Method & L-Thorax & L-Shoulder & L-Elbow & L-Wrist & L-Hand \\ 
        \midrule
        PHC+ & 45.0 & 45.5 & 43.3 & 40.3 & 41.7\\
        \name & \textbf{20.6} & \textbf{21.4} & \textbf{27.0} & \textbf{29.3} & \textbf{31.5}\\
        \bottomrule
    \end{tabular}
    \label{tab:left_upper_limb}
\end{table} 
\section{Joint-wise Tracking Error}
\label{sec:error}
Full-body tracking requires high precision. We present comprehensive joint-wise tracking error statistics across the five body parts defined in our skill modularization framework: right arm (\cref{tab:right_upper_limb}), left arm (\cref{tab:left_upper_limb}), right leg (\cref{tab:right_lower_limb}), left leg (\cref{tab:left_lower_limb}), and torso (\cref{tab:core_and_pelvis}). These detailed analyses highlight the effectiveness of skill modularization in achieving precise, high-fidelity physical character skill learning, demonstrating its potential for fine-grained motion control across distinct body parts.\\

\noindent Additionally, we observe higher errors at the end-effector joints (e.g., R-Toe, L-Toe, R-Hand, L-Hand) for both \name and PHC+, suggesting that more complex motion samples or refined modeling of these joints may be needed. We also note increased errors in the right leg joints for our method compared to other body parts, which may indicate a bias or imbalance in the training data. However, the modular nature of our framework offers key advantages: we can not only isolate part-specific controllers during training but also leverage active learning to generate additional synthesized samples, helping to support skill learning for specific body parts.\\

\begin{table}[h]
    \centering
    \footnotesize
    \caption{\textbf{L-Leg:} Joint-wise Mean Position Errors (mm) on AMASS-Test.}
    \begin{tabular}{@{}l|cccc@{}}
        \toprule
        Method & L-Hip & L-Knee & L-Ankle & L-Toe \\ 
        \midrule
        PHC+ & 38.7 & 43.4 & 52.5 & 55.7\\
        \name & \textbf{20.1} & \textbf{26.3} & \textbf{31.7} & \textbf{33.9}\\
        \bottomrule
    \end{tabular}
    \label{tab:left_lower_limb}
\end{table}

\begin{table}[h]
    \centering
    \footnotesize
    \caption{\textbf{R-Leg:} Joint-wise Mean Position Errors (mm) on AMASS-Test.}
    \begin{tabular}{@{}l|cccc@{}}
        \toprule
        Method & R-Hip & R-Knee & R-Ankle & R-Toe \\ 
        \midrule
        PHC+ & 36.4 & 41.4 & 50.2 & 50.9\\
        \name & \textbf{21.7} & \textbf{31.6} & \textbf{38.9} & \textbf{47.9}\\
        \bottomrule
    \end{tabular}
    \label{tab:right_lower_limb}
\end{table}

\begin{table}[h]
    \centering
    \footnotesize
    \caption{\textbf{Torso:} Joint-wise Mean Position Errors (mm) on AMASS-Test.}
    \begin{tabular}{@{}l|cccccc@{}}
        \toprule
        Method & Pelvis & Torso & Spine & Chest & Neck & Head \\ 
        \midrule
        PHC+ & 36.9 & 39.1 & 43.2 & 43.8 & 44.2 & 48.5\\
        \name & \textbf{19.7} & \textbf{21.0} & \textbf{22.0} & \textbf{21.6} & \textbf{20.0} & \textbf{20.5}\\
        \bottomrule
    \end{tabular}
    \label{tab:core_and_pelvis}
\end{table}

\begin{figure}[h]
    \centering
    \includegraphics[width=\linewidth]{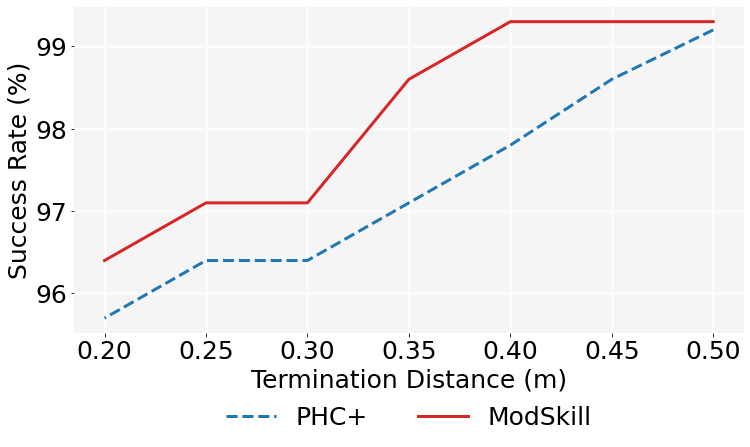}
    \vspace{-0.5cm}
    \caption{Effect of decreasing termination distance on full-body motion tracking success rate for AMASS-Test.}
    \label{fig:terminate}
\end{figure}

\noindent Following prior work \cite{Luo2023PerpetualHC}, we use early termination during training and evaluation, where tracking terminated (i.e. considered a failure) if the average deviation of simulated joint positions from the reference exceeds 0.5 meters. However, this threshold is relatively lenient, as both the use of average deviation and the large 0.5-meter margin may overlook significant joint-wise errors that aren't fully captured by the current evaluation metrics. In Fig. \ref{fig:terminate}, we report the full-body tracking success rate on AMASS-Test for termination distances ranging from 0.5 meters to 0.2 meters, with a step size of 0.05 meters. Notably, we observe an immediate decline in performance for PHC+ as the termination distance decreases, whereas our method maintains a higher success rate even under stricter termination conditions. This suggests that our approach is more robust, able to generalize effectively, and consistently preserves higher accuracy across a wider range of unseen sequences.\\

\begin{figure*}
    \centering
    \includegraphics[width=\linewidth]{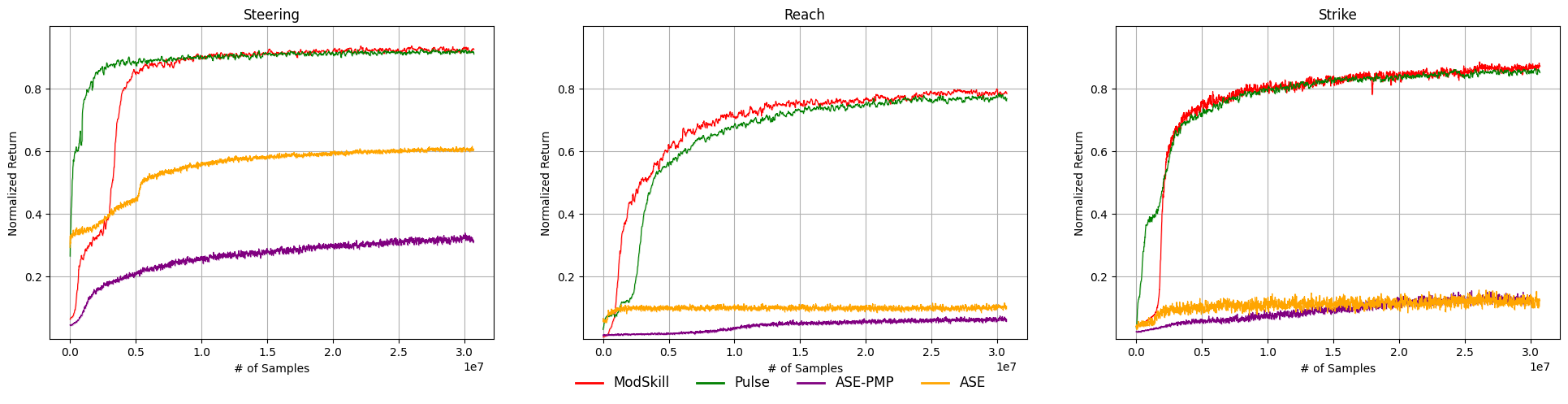}
    \caption{Normalized returns during training for downstream tasks.}
    \label{fig:transfer}
\end{figure*}
\begin{figure*}[h]
    \centering
    \includegraphics[width=0.9\linewidth]{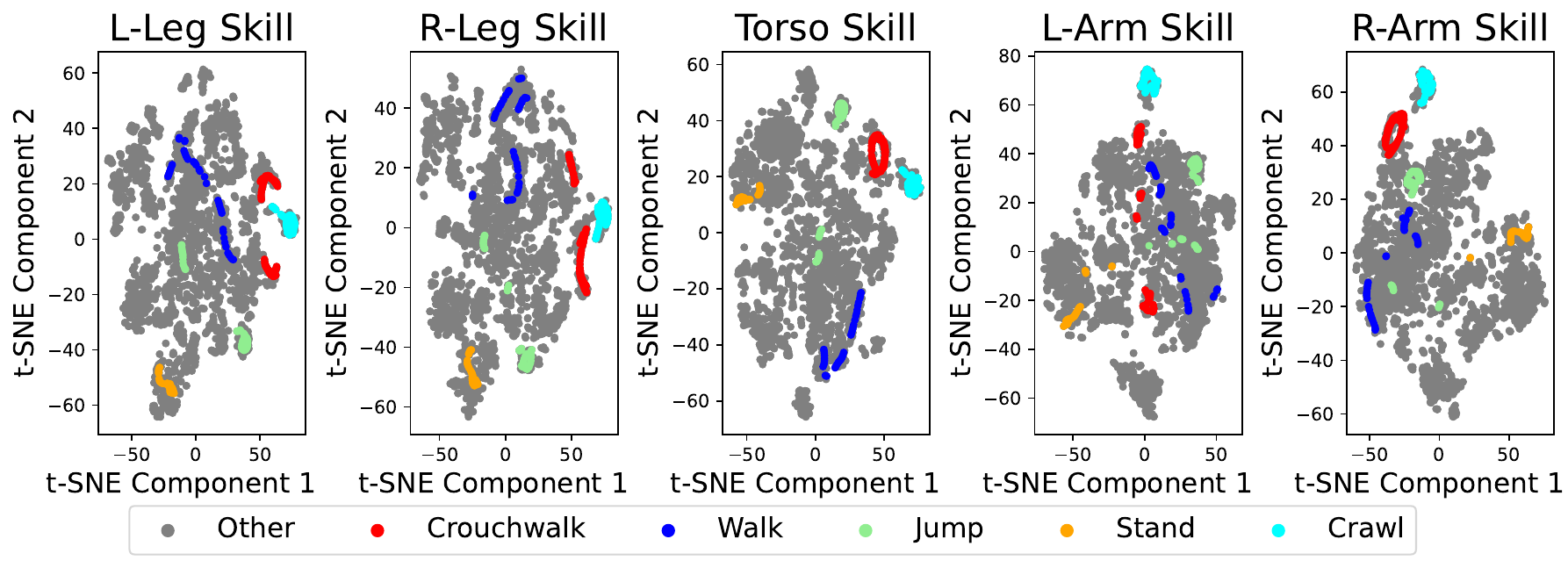}
    \caption{Modular Skill Embedding t-SNE Visualization.}
    \label{fig:tsne}
\end{figure*}
\section{Downstream Tasks}
\label{sec:gen}
\noindent For generative downstream tasks, steer, reach, strike, and VR-tracking, we utilize a three-layer MLP with dimensions [2048, 1024, 512] for the high-level policy and train the policy using PPO \cite{schulman2017proximalpolicyoptimizationalgorithms} for approximately $2 \times 10^9$ steps with a learning rate of $2 \times 10^{-5}$. VR-tracking follows the reward function for the full-body tracking tasks. Please refer to our supplementary video for a comprehensive evaluation of our model on these tasks. Following \cite{2022-TOG-ASE, luo2024universal}, the goal state and reward formulations for steer, reach, and strike are detailed below:\\ 

\paragraph{Steering.} The goal state is defined as \( s_{\text{steer}}^t := (d_t, v_t) \), where \( d_t \) and \( v_t \)  represent the target direction and the desired linear velocity at timestep \( t \), respectively. The objective of the policy is to control the character to travel along the specified direction at the desired velocity. The reward is defined as \( r_{\text{speed}} = |v_t - v_{t0}| \), where \( v_{t}^{0} \) is the root velocity of the humanoid.

\paragraph{Reach.} In the reach task, we aim to minimize the distance between the simulated character's right hand and a desired 3D target point, \( c_t \), randomly sampled from a 2-meter box centered at \( (0, 0, 1) \). The goal state is \( s_{\text{reach}}^t \equiv (c_t) \). Let \( p_{\text{R-Hand}} \) be the position of the simulated character's right hand. The reward for reaching is calculated as the exponential of the negative squared distance between the right-hand position and the desired target point: 
\[
r_{\text{reach}} = \exp\left(-5 \| p_{\text{R-Hand}} - c_t \|_2^2 \right),
\]

\paragraph{Strike.} The objective of this task is to knock over a target object. We select the rigid bodies, R-Hand, R-Wrist, R-Elbow as the target body parts for contact, where the task terminates (e.g. considered a failure) if any body part other than the target body parts makes contact with the target object. The goal state \( s_{\text{strike}}^t \equiv (x_t, \dot{x}_t) \) consists of the position and orientation \( x_t \), linear and angular velocities \( \dot{x}_t \) of the target object in the simulated character's frame of reference. The reward is defined as \( r_{\text{strike}} = 1 - \mathbf{u}_{\text{up}} \cdot \mathbf{u}_t \), where \( \mathbf{u}_{\text{up}} \) is the global up vector, and \( \mathbf{u}_t \) is the up vector of the target object.\\

\noindent As shown in Fig. \ref{fig:transfer}, we provide the training curves for the steering, reach, and strike tasks. We observe that our method initially exhibits slower progress than SOTA models, especially for tasks that utilize full-body template movements, such as speed and strike, due to the added complexity of modular skill spaces. However, the learning curve accelerates quickly, ultimately catching up to SOTA models with a similar convergence speed and scale of normalized returns. For more precise tasks, like reaching, that require targeted control of specific body parts, our method demonstrates faster learning, indicating better flexibility for character control. 

\section{Modular Skill Embedding Space}
\label{sec:viz}
In Fig. \ref{fig:tsne}, we present the t-SNE visualization of body-part skill embeddings for motions from AMASS-Test. We uniformly sample skill embeddings for each body part every 10 frames, resulting in $\sim$3000 samples. For clarity, we label a subset of samples for five different types of motions. We observe consistent structures for the same motion across body parts, with symmetric structures emerging for motions with alternating patterns, e.g., walking and jumping.


\end{document}